\documentclass{article}

\PassOptionsToPackage{numbers, compress}{natbib}
\pdfoutput=1
\usepackage[final]{nips_2017}

\usepackage[utf8]{inputenc} 
\usepackage[T1]{fontenc}    
\usepackage{hyperref}       
\usepackage{url}            
\usepackage{booktabs}       
\usepackage{amsfonts}       
\usepackage{nicefrac}       
\usepackage{microtype}      
\usepackage{subfigure}
\usepackage[labelfont=bf,font=normalsize]{caption}
\usepackage[makeroom]{cancel}

\usepackage{amsmath,amsthm}
\usepackage{amsfonts}
\usepackage{amssymb}
\usepackage{enumerate}
\usepackage{float}
\usepackage{bm}
\usepackage{graphicx}
\usepackage{natbib}
\usepackage{multirow}
\usepackage{dirtytalk}
\usepackage{array}

\usepackage{algorithm}
\usepackage{algorithmic}

\newcommand{\comment}[1]{}

\usepackage{xspace} 

\newcommand{\ogan}{$\beta'$-GAN\xspace}
\newcommand{\fequ}{\emph{favorite equilibrium}\xspace}
\newcommand{\tgan}{\textit{tiny}-GAN\xspace}
\newcommand*\diff{\mathop{}\!\mathrm{d}}

\DeclareMathOperator*{\argmin}{arg\,min}
\DeclareMathOperator*{\expectation}{\mathbb{E}}

\title{Analysis of Nonautonomous Adversarial Systems}

\author{
  Arash Mehrjou\\
  Max Planck Society\\
  \texttt{amehrjou@tuebingen.mpg.de} \\
}

\begin{document}

\maketitle

\begin{abstract}
Generative adversarial networks are used to generate images but still their convergence properties are not well understood. There have been a few studies who intended to investigate the stability properties of GANs as a dynamical system. This short writing can be seen in that direction. Among the proposed methods for stabilizing training of GANs, $\beta$-GAN was the first who proposed a complete annealing strategy to change high-level conditions of the GAN objective. In this note, we show by a simple example how annealing strategy works in GANs. The theoretical analysis is supported by simple simulations.
\end{abstract}

\section{Introduction}
Generative adversarial nets~\citep{goodfellow2014generative} are trained by optimizing an objective function over two sets of parameters $\{\theta, \psi\}$. For ease of presentation, the framework is described as a competition between two functions, generator and discriminator who want to minimize/maximize a mutual objective. The GAN objective in its general shape can be written as
\begin{equation}
\argmin_{\theta}\max_{\psi}\mathcal{L}(\theta,\psi)=\expectation_{p(z)}[f(D_{\psi}(G_{\theta}(z)))]+\expectation_{p_D(x)}[f(-D_{\psi}(x))]
\label{eq:fullGAN_objective}
\end{equation}
Where $\psi$ parameterizes the discriminator and $\theta$ parameterizes the generator. Different choices for $f(.)$ gives various GAN objectives, e.g. Jenson-Shannon~\citep{goodfellow2014generative}, Wassestein-GAN~\citep{arjovsky2017wasserstein}, f-GAN~\citep{nowozin2016f}, etc. In accordance with these works, we assume $f'(x)\neq 0$. The ultimate goal is to converge to a saddle point where neither discriminator nor generator can achieve a better objective when the other one is kept fixed. Let's call this point in the $(\theta, \psi)$ space the \fequ. The interesting property of this point is that $G_{\theta}(z)=p_D(x)$. Currently, people are using stochastic gradient descent (SGD) updates to alternately perturb $\theta$ and $\psi$ in a hope to converge to the \fequ in the end. Even though the results look visually promising, the dynamical behavior of this system needs more investigation.

In this note, we restrict ourselves to a minimal example and try to get some insight of a method called \emph{Annealed Generative Adversarial Networks} (a.k.a $\beta$-GAN) which was proposed last year and proved to be effective in stabilizing the optimization of GANs in practice.

\section{Nonautonomous GAN}
{\it Continuous dynamical system---} We see GAN as a continuous dynamical system. This assumption is valid when the learning rate of Stochastic Gradient Descent (SGD) tends to zero in optimization, i.e. $\epsilon \to 0$.

{\it Autonomous GAN ---}In conventional GAN training, the dynamical system

\begin{equation}
\left\{
	\begin{array}{ll}
		\dot{\theta}=-\nabla_{\theta}\mathcal{L}(\theta,\psi)\\
        \dot{\psi}=\nabla_{\psi}\mathcal{L}(\theta,\psi)
	\end{array}
    \label{eq:continuous_updates}
\right.
\end{equation}

is an approximation of the training pattern for a tiny learning rate. We call these dynamical systems \emph{autonomous} because the right-hand side function is not an explicit function of time~\citep{khalil1996noninear}. Given the Lipschitz continuity of the right-hand side of Eq.~\ref{eq:continuous_updates}, there exists a solution for this system and it is unique.

{\it Nonautonomous GAN---}
The overall idea is introducing a new state $\alpha$ in the GAN objective function in Eq.~\ref{eq:fullGAN_objective}. This state controls the data distribution. More precisely, the objective function becomes
\begin{equation}
\mathcal{L}(\theta,\psi,\alpha)=\expectation_{p(z)}[f(D_{\psi}(G_{\theta}(z)))]+\expectation_{p_D(x;\alpha)}[f(-D_{\psi}(x))]
\label{eq:nonautoGAN_objective}
\end{equation}
To study the effect of this new state, we introduce a minimalistic framework called  \tgan to emphasize our points: 

{\it tiny-GAN ---}To have a minimal tractable GAN framework, we set $P_D(x;\alpha)=\delta_{\alpha}$ and $G_{\theta}(z)=\delta_{\theta}$, meaning that real data is concentrated on a single point at $x=\alpha$ and the generator is only capable of generating one point at location $x=\theta$. The discriminator is assumed linear, i.e. $D_{\psi}(x)=\psi x$. In contrast to ~\citep{mescheder2018convergence}, we do not tie data to the origin and release it to occupy any location on the real axis. After these simplifications, the objective function of Eq.~\ref{eq:fullGAN_objective} becomes:
\begin{equation}
\mathcal{L}(\theta,\psi,\alpha) = f(\psi\theta)+f(-\psi\alpha)
\end{equation}
and the dynamical system of training GAN in Eq.~\ref{eq:continuous_updates} is written as
\begin{equation}
\left\{
	\begin{array}{ll}
		\dot{\theta}=-\psi f'(\psi\theta)\\
        \dot{\psi}=\theta f'(\psi\theta) -\alpha_r f'(-\psi\alpha_r)\\
	\end{array}
    \label{eq:auto_continuous_updates}
\right.
\end{equation}
In this formulation, $\alpha_r$ is fixed and represents real data distribution.

\begin{figure}[t!]
	\centering
\subfigure[Autonomous GAN]{
\includegraphics[width=0.42\linewidth]{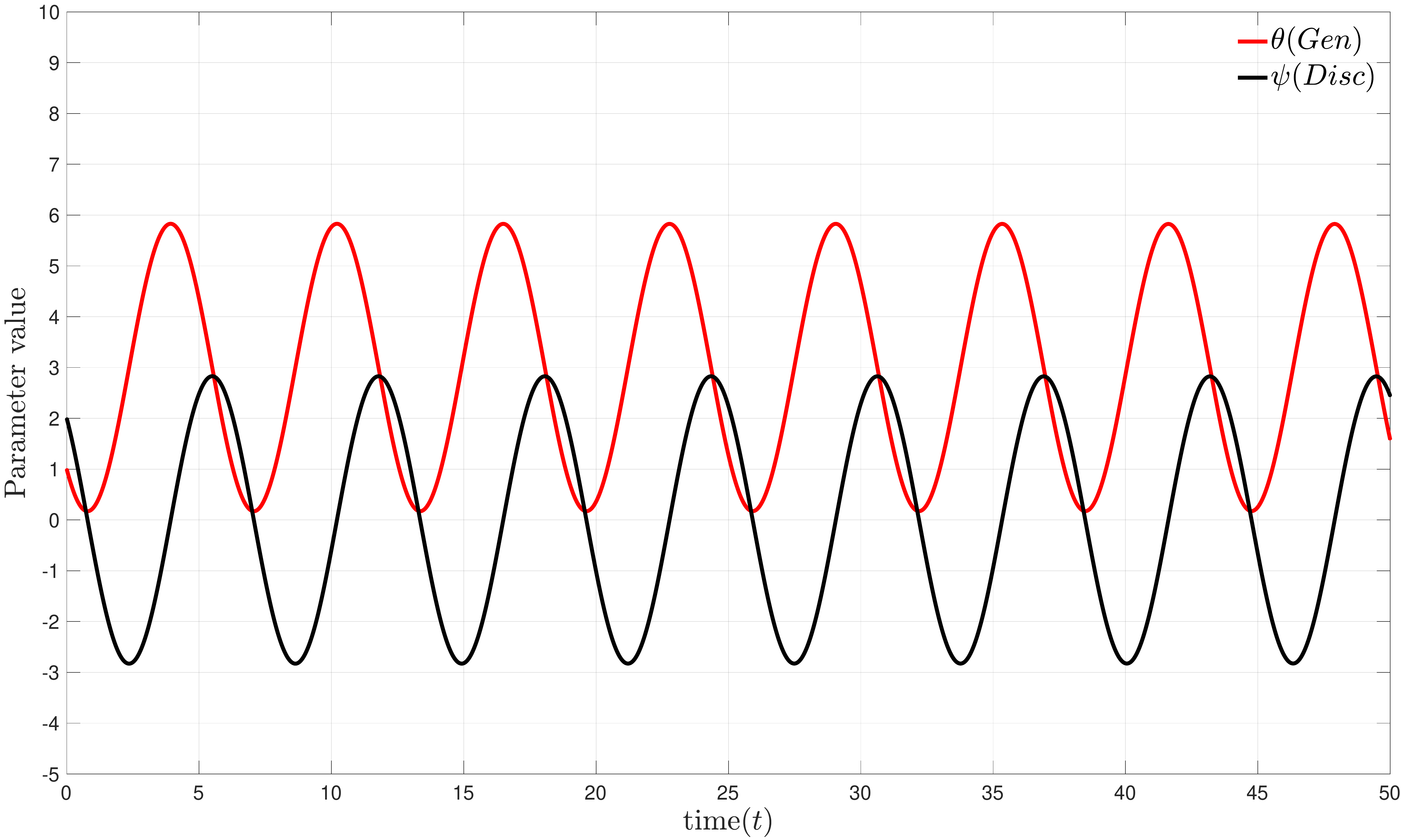}}
\subfigure[$\beta'$-GAN: coupled-dynamics]{
\includegraphics[width=0.42\linewidth]{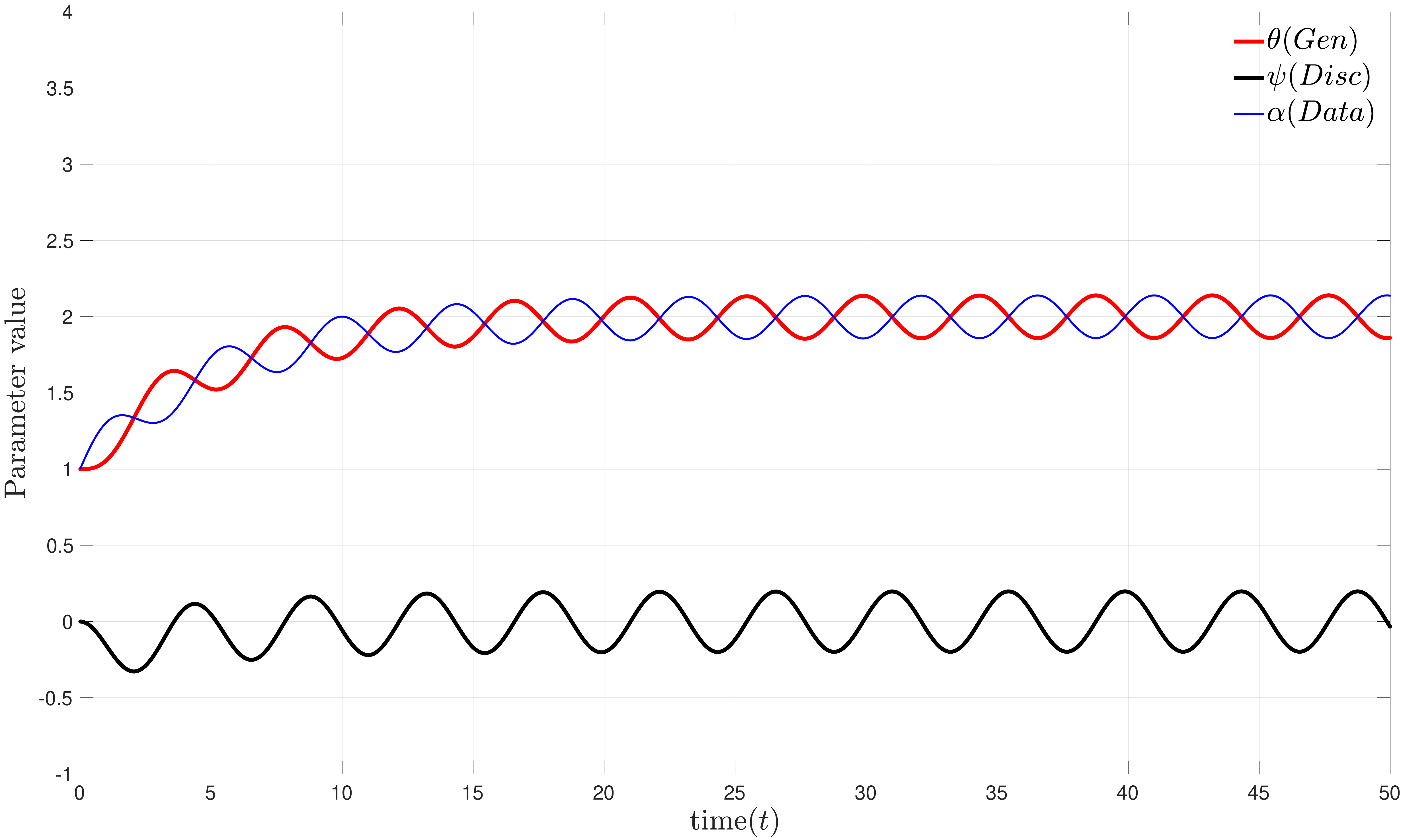}}
\subfigure[$\beta$-GAN: decoupled-dynamics]{
\includegraphics[width=0.42\linewidth]{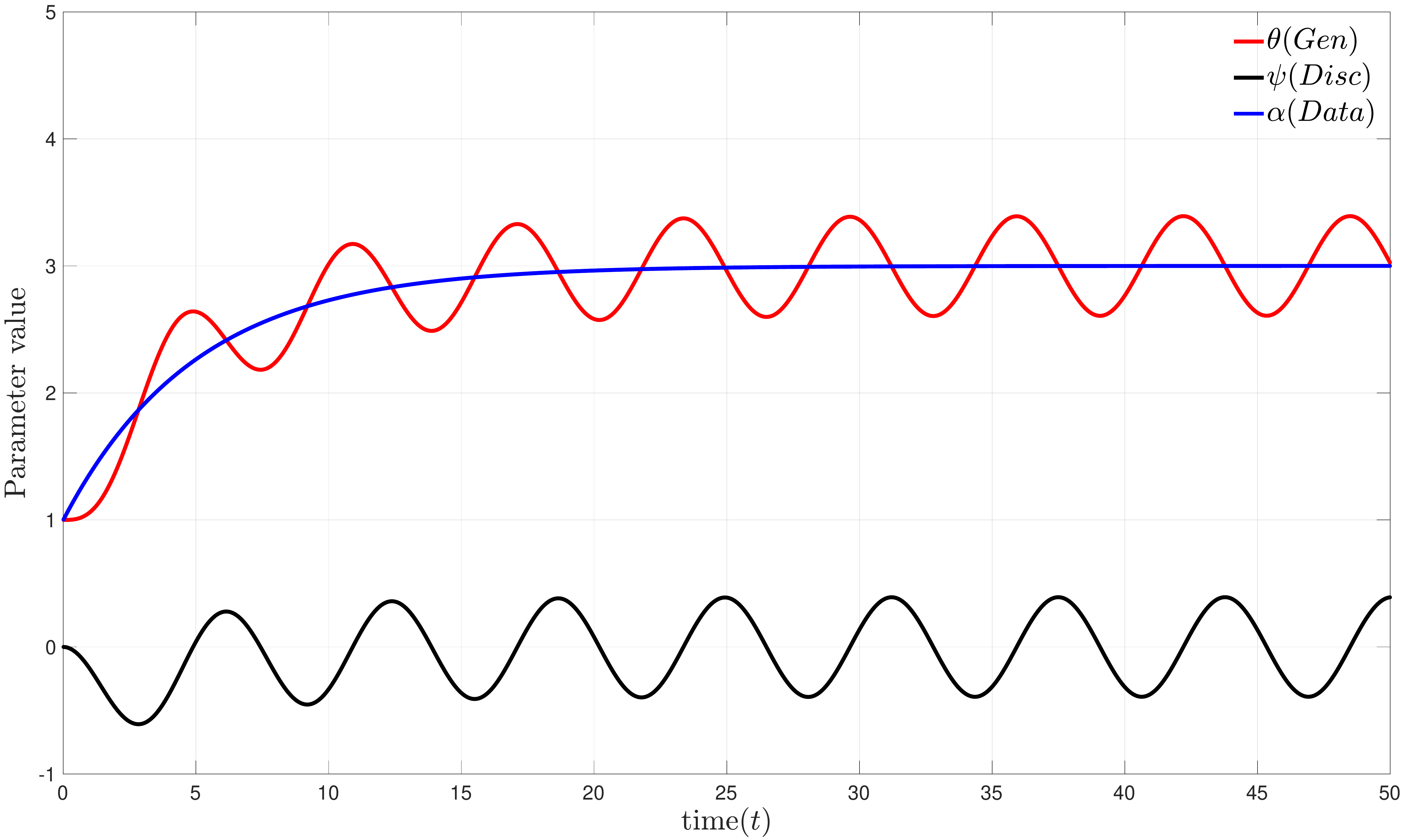}}
\subfigure[$\beta$-GAN: decoupled-dynamics (long T)]{
\includegraphics[width=0.42\linewidth]{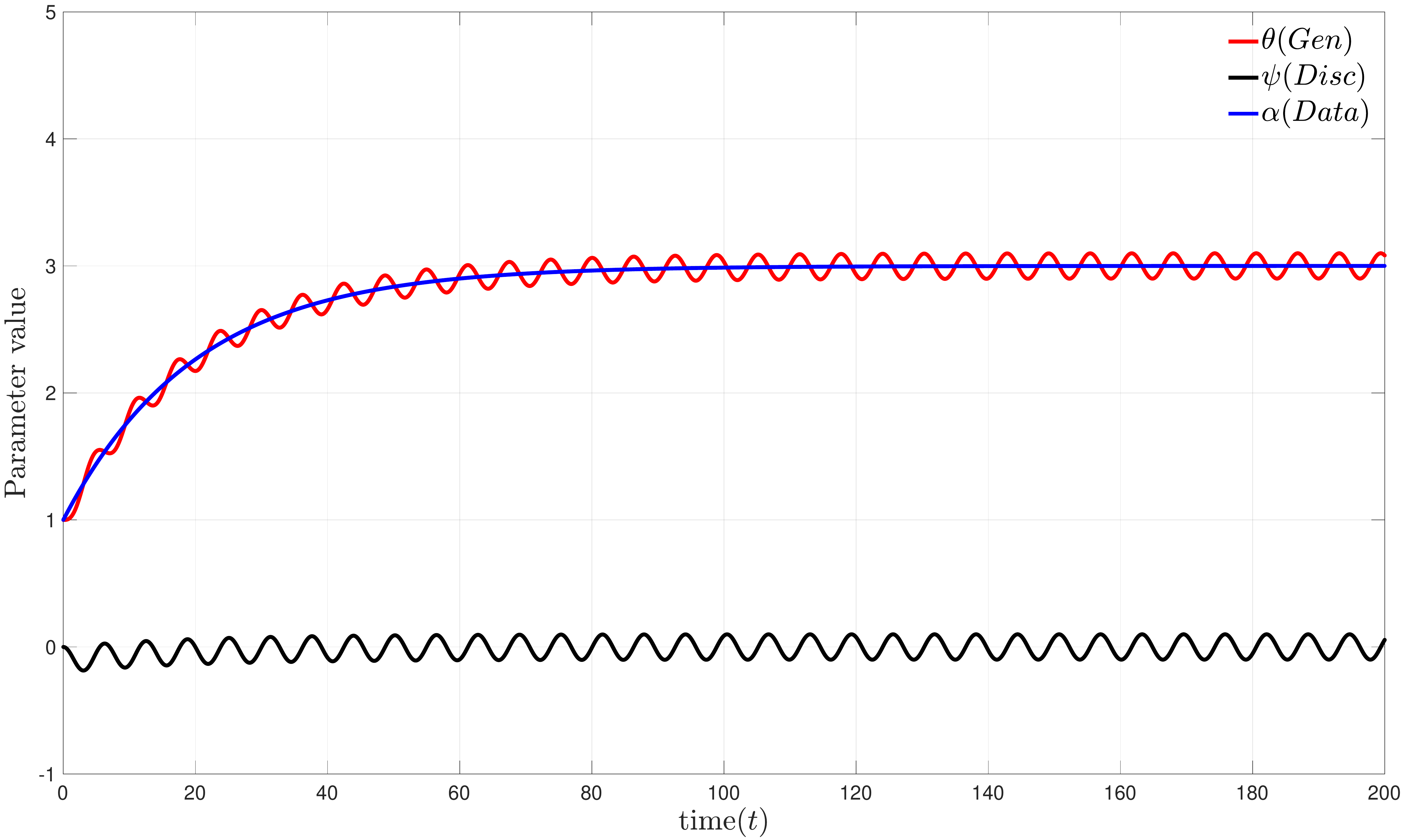}}
	\caption{State evolution of various continuous dynamical systems approximating the behavior of GANs when the learning rate is small $\epsilon\to 0$.(a) Two-state autonomous GAN with static data distribution. (b) Three-state Nonautonomous GAN when the dynamics of data distribution is coupled with the other states. (c) Three-state Nonautonomous GAN when the dynamics of data distribution is only governed by the annealing process. (d) The same as (c) but with slower annealing process.}
	\label{fig:nonautonomous-GAN}
	\vspace{-0.2cm}
\label{fig:simulations}
\end{figure}

Many formulations of GAN can be characterized by the dynamical system of Eq.~\ref{eq:continuous_updates} which contains only two states: the parameters of the generator ($\theta$) and the parameters of the discriminator ($\psi$). Here we augment the state-space equation with a new state which characterizes the properties of the data distribution $p_D(x)$. In harmony with the minimalistic nature of \tgan, the entire data distribution is characterized by $\alpha$ here. Notice that the real data distribution is not dynamic. Indeed, real data distribution is the target point of the dynamics of $\alpha(t)$ and we represent it by $\alpha_r$, i.e. $\alpha(t)\to \alpha_r$ as $t\to\infty$.
Optimizing Eq.~\ref{eq:nonautoGAN_objective} when the dynamics of $\alpha$ is only governed by $\nabla_\alpha\mathcal{L}(\theta,\psi,\alpha)$ results in trivial answers since there will be no guarantee that $\alpha(t)$ and $\theta(t)$ arrives at \fequ where $\alpha(\infty)=\alpha_r$. To cure this issue, $\beta$-GAN suggested a full annealing strategy over $p_D(x;\alpha)$. This idea turns the dynamical system of Eq.~\ref{eq:continuous_updates} into a time-varying (Nonautonomous) system. At this point two branches can be thought of. In the first branch which is also the method devised by $\beta$-GAN, $\alpha$ has partially decoupled dynamics from the other states of the system. By partially decoupled, we mean that the dynamics of $\alpha$ is not affected by the dynamics of the other states of the system. However, the dynamics of the other states may depend on the dynamics of $\alpha$. In the second branch (let's call it \ogan), $\alpha$ undergoes two dynamics. One is the dynamics imposed by the GAN objective $\nabla_\alpha\mathcal{L}(\theta,\psi,\alpha)$ which acts by SGD updates and the other one is the annealing dynamics. The first term makes the dynamics of $\alpha$ coupled with the other states of the system. As proposed in $\beta$-GAN, annealing steps must act with a slower timescale than SGD iterations of the optimization. The slow partially decoupled dynamics of $\alpha$ is characterized by 
\begin{equation}
\alpha(t)=(\alpha_0-\alpha_r)e^{-\frac{t}{T}}+\alpha_r
\end{equation}
where $T>1$ is a time constant that makes this dynamic term slower than the SGD dynamics. In addition, $\alpha_0$ is the initial value of $\alpha(t)$ that characterizes the initial distribution of data $p_D(x;\alpha=\alpha_0)$ when the annealing process starts and $\alpha_r$ is the target value of $\alpha(t)$ for which $p_D(x;\alpha=\alpha_r)$ becomes the real data distribution $p_D(x)$. Therefore the state-space equation is written as follows:
\begin{equation}
\left\{
	\begin{array}{ll}
		\dot{\theta}=-\psi f'(\psi\theta)\\
        \dot{\psi}=\theta f'(\psi\theta) -\alpha f'(-\psi\alpha)\\
        \dot{\alpha}=\lambda[-\psi f'(-\psi\alpha)]+\frac{1}{T}(\alpha_r - \alpha_0)e^{-\frac{t}{T}}
	\end{array}
    \label{eq:nonauto_continuous_updates}
\right.
\end{equation}
The hyper-parameter $\lambda\in\{0,1\}$ is a switch and has an important meaning which differentiate between $\beta$-GAN and \ogan. When $\lambda=0$ ($\beta$-GAN) the variable $\alpha$ is not perturbed by short timescale SGD updates. This means that $\alpha$ has partially decoupled dynamics from the dynamics of states $\{\theta, \psi\}$. On the other hand, when $\lambda=1$, the dynamics of $\alpha$ is governed by both a short timescale term and a long timescale term. The former is the SGD updates and the latter is the same as in $\beta$-GAN. Furthermore, $\beta$-GAN suggests starting from \emph{uniform} distribution meaning that $p_D(x;\alpha)$ is constant over a specified area and zero elsewhere. The generator must be pre-trained to capture the uniform distribution for a certain data dimension $n$. This means that the generator at time $t=0$ is able to generate a simple uniform distribution which matches the initial distribution $p_D(x;\alpha=\alpha_0)$. In our minimalistic setting of \tgan and the dynamical system of Eq.\ref{eq:nonauto_continuous_updates}, this translates to $\theta(0)=\alpha(0)=\alpha_0$.

\section{Simulations}
\label{sec:simulatins}
To show the effect of annealing strategy in GANs, simple simulations are presented here for autonomous GAN, $\beta$-GAN and \ogan. Note that the objective function of Eq.~\ref{eq:fullGAN_objective} becomes that of W-GAN when $f(y)=y$. We compare normal (autonomous) GAN  with two Nonautonomous GANs ($\beta$-GAN and \ogan). Remember that in $\beta$-GAN, data distribution does not change with short timescale and it has its own partially decoupled dynamics due to annealing while in \ogan, data distribution is altered by both the fast dynamics of SGD and the slow dynamics of annealing. In all simulated experiments, the real data distribution is located at $\alpha_r=3$ which is the static value of $\alpha$ for autonomous GAN but target value of $\alpha(t)$ for Nonautonomous GANs. Fig.~\ref{fig:simulations}(a) shows the solution of the dynamical system of Eq.~\ref{eq:auto_continuous_updates} when $f(y)=y$ as in Wasserstein GAN with initial point $(\theta(0),\psi(0))=(1,2)$. As can be seen, the states $(\theta,\psi)$ are oscillating around $(\theta^*, \psi^*)$=$(3, 0)$ which is the equilibrium point of this system. For the linear $f(y)=y$ and \tgan framework used in the note, this result is global. It can be shown that for nonlinear choices of $f(y)$, the same oscillation is observable but locally around the equilibrium point. Notice that this oscillation is so called \emph{unsustained oscillation} which is different from stable limit cycles. Here, the amplitude of the oscillation depends on the initial state $(\theta(0), \psi(0))$ which is an undesirable effect.
Fig.~\ref{fig:simulations}(b) depicts the behavior of $\beta'$-GAN and shows the solution to the dynamical system of Eq.~\ref{eq:nonauto_continuous_updates} when $\lambda=1$ with initial states $(\theta(0),\psi(0),\alpha(0))=(1,0,1)$. Still the target value for $\alpha(t)$ is $\alpha_r=3$. As can be seen, the dynamical system is still oscillating but the amplitude of the oscillation is reduced. The bad thing here is that the system is oscillating around a wrong point $(\theta,\psi,\alpha)=(2,0,2)$ which is different from the \emph{favorite equilibrium} $(\theta,\psi,\alpha)=(3,0,3)$. 

Fig.~\ref{fig:simulations}(c) simulates the behavior of $\beta$-GAN by running the dynamical system of Eq.~\ref{eq:nonauto_continuous_updates} with $\lambda=0$ from the initial states $(\theta(0),\psi(0),\alpha(0))=(1,0,1)$. Again the target data is $\alpha_r=3$ and $T=3$. As can be seen, the system is oscillating as Fig.~\ref{fig:simulations}(b) but this time around the correct point $(\theta,\psi,\alpha)=(3,0,3)$. The amplitude of oscillation is lower than autonomous GAN of Fig.~\ref{fig:simulations}(a) and decreases more by increasing $T$. Increasing $T$ means it takes longer for $\alpha$ to move from $\alpha_0$ to $\alpha_r$ which is equivalent to slower annealing dynamics or finer annealing steps in discrete setting. This is shown in Fig.~\ref{fig:simulations}(d) where the entire setting is as the previous case but $T=30$ results in slower approach to $a_r$ but reduced oscillation amplitude around the correct equilibrium point.

\section{Theoretical Analysis}
The simulations of section~\ref{sec:simulatins} shows that the amplitude of oscillation decreases as $T$ increases in $\beta$-GAN framework. Here, a more formal analysis is provided to explain this observation. The dynamical system of Eq.~\ref{eq:nonauto_continuous_updates} for $f(y)=y$ and $\lambda=0$ will be written as follows:

\begin{equation}
\left\{
	\begin{array}{ll}
		\dot{\theta}&=-\psi\\
        \dot{\psi}&=\theta-\alpha\\
        \dot{\alpha}&=\frac{1}{T}(\alpha_r - \alpha_0)e^{-\frac{t}{T}}.
	\end{array}
  \label{eq:simple_nonauto_continuous_updates}
\right.
\end{equation}
Let's $a=\frac{1}{T}$ and $K=\frac{(\alpha_r-\alpha(0))}{T}$. We take Laplace transform from both sides of three equations above:
\begin{equation}
\left\{
	\begin{array}{ll}
		s\theta(s)-\theta(0)&=-\psi(s)\\
        s\psi(s)-\psi(0)&=\theta(s)-\alpha(s)\\
        s\alpha(s)-\alpha(0)&=\frac{K}{s+a}.
	\end{array}
  \label{eq:laplace_nonauto_continuous_updates}
\right.
\end{equation}
Taking derivative of the both sides of the second line of Eq.~\ref{eq:simple_nonauto_continuous_updates} amounts to multiplying both sides of the second line of Eq.~\ref{eq:laplace_nonauto_continuous_updates} by Laplace differentiation operator $s$ and results in

\begin{equation} 
s^2\psi(s)=s\theta(s)-s\alpha(s)
        =\cancel{\theta(0)}-\psi(s) - \cancel{\alpha(0)}-\frac{K}{s+a}
\end{equation}

where $\theta(0)$ and $\alpha(0)$ cancels each other due to the assumption of $\beta$-GAN that generator starts from a simple initial distribution characterized by $\alpha(0)$. This assumption consequently ensures $\psi(0)=0$ because it is assumed that the the equilibrium is initially found for both generator and discriminator for the data distribution $\alpha(0)$. Solving for $\psi(s)$ gives us
\begin{equation}
\psi(s)=\frac{-K}{(1+s^2)(s+a)}.
\end{equation}
We then expand the right-hand side as a sum of polynomial fractions:
\begin{equation}
\psi(s)=\frac{-K}{1+a^2}\frac{s}{1+s^2}+\frac{Ka}{1+a^2}\frac{1}{1+s^2}+\frac{K}{1+a^2}\frac{1}{s+a}.
\end{equation}
Computing inverse Laplace transform of $\psi(s)$ gives 
\begin{equation}
\mathcal{L}^{-1}\{\psi(s)\} = \overbrace{A\cos(t)+B\sin(t)}^{\psi_1(t)}+Ce^{-at}. 
\end{equation}
where $A=\frac{-K}{1+a^2}$, $B=\frac{Ka}{1+a^2}$, and $C=\frac{K}{1+a^2}$. The last term vanishes in the steady state solution when $t\to\infty$. We are mainly interested in the first two parts which are responsible for the persistent oscillation. Adding two harmonics results in a new harmonic with scaled amplitude $\mathcal{A}$ and phase shift $\phi$:
\begin{equation}
\left\{
	\begin{array}{ll}
   		 \psi_1(t)=\mathcal{A}\sin(t+\phi)\\
		\mathcal{A}=\sqrt{A^2+B^2+2AB\cos(\pi/2)}\\
        \phi=\tan^{-1}(A,B)
	\end{array}
\right.
\end{equation}
where $\tan^{-1}$ is quadrant-aware arc tangent. By substituting $A$ and $B$ in $\mathcal{A}$ we can compute the amplitude of the persistent oscillation as
\begin{equation}
\mathcal{A}=\sqrt{\left(\frac{K}{1+a^2}\right)^2+\left(\frac{Ka}{1+a^2}\right)^2}=\frac{K}{1+a^2}\sqrt{(1+a^2)}.
\end{equation}
The term $1+a^2=1+\frac{1}{T^2}\to 1$ as $T\to\infty$. The important term is $K$ that goes to zero as $T\to\infty$ and proves our point that the amplitude decreases as the annealing time $T$ increases.
Now that the analytic form of $\psi(t)$ is known, we can move on and obtain the analytic form of $\theta(t)$. According to Eq.\ref{eq:laplace_nonauto_continuous_updates}, we can write $\theta(s)$ in terms of $\psi(s)$ as follows:
\begin{equation}
\theta(s)=\frac{1}{s}[\theta(0)-\psi(s)]
\label{eq:theta_laplace}
\end{equation}
where $\frac{1}{s}$ acts as an integrator. Therefore, we can obtain the inverse Laplace transform $\mathcal{L}^{-1}\{\theta(s)\}$ and compute the following definite integral to compute $\theta(t)$ as

\begin{align}
\theta(t)&=\theta(0)\mathcal{L}^{-1}\{\frac{1}{s}\}-\int_{\tau=0}^{\tau=t}\psi(\tau)\diff \tau\nonumber\\
&=\theta(0)-\frac{K}{1+a^2}\int_{\tau=0}^{\tau=t}e^{-at}\diff \tau+\int_{\tau=0}^{\tau=t}\psi_1(\tau)\diff \tau\nonumber\\
&= \theta(0)+\frac{K}{1+a^2}\frac{1}{a}+\int_{\tau=0}^{\tau=t}\psi_1(\tau)\diff \tau\nonumber\\
&= \theta(0)+\frac{\alpha_r-\alpha_0}{1+a^2}+\underbrace{\int_{\tau=0}^{\tau=t}\psi_1(\tau)\diff \tau}_{\Psi_1(t)}.
\end{align}
Notice that $\Psi_1(t)$ is the integral of a sinusoidal which is itself a sinusoidal. As the annealing time increases, $T\to\infty$, the term $1+a^2=1+\frac{1}{T^2}\to1$ and we eventually have the steady state solution of $\theta(t)$ as follows:
\begin{equation}
\lim_{T\to\infty}\theta(t)=a_r+\Psi_1(t)
\end{equation}
which shows the oscillation around the desired equilibrium point $a_r$ that is the real data distribution. \qed

\section{Conclusion}
This writing suggests annealing as a promising approach in GANs. The practical results are already provided in $\beta$-GAN paper~\citep{mehrjou2017annealed}. In this note, a minimalistic nonautonomous adversarial system is proposed to mimic the behavior of GAN in a tractable way when its data distribution is changing. The optimization updates and the dynamics of the annealing strategy is approximated by a continuous dynamical system. Simulations and theoretical analysis are performed to give insight into the dynamics of GANs under annealing. We believe viewing adversarial strategies as dynamical systems are interesting not only in unsupervised learning, but also in control theory where compelling systems may arise when states act in an adversarial way.

\bibliographystyle{plain}
\bibliography{main}

\end{document}